%% file: main.tex
\documentclass[10pt,twocolumn,letterpaper]{article}

\usepackage[pagenumbers]{cvpr} 
\usepackage{graphicx}
\usepackage{amsmath}
\usepackage{amssymb}
\usepackage{booktabs}
\usepackage{multicol}

\usepackage[pagebackref,breaklinks,colorlinks]{hyperref}

\usepackage[capitalize]{cleveref}
\crefname{section}{Sec.}{Secs.}
\Crefname{section}{Section}{Sections}
\Crefname{table}{Table}{Tables}
\crefname{table}{Tab.}{Tabs.}

\usepackage{tablefootnote}
\usepackage{wrapfig}
\usepackage{tabularx,ragged2e}
\newcolumntype{Y}{>{\centering\arraybackslash}X}
\newcolumntype{Z}{>{\centering\arraybackslash}p{0.15\textwidth}X}
\usepackage{adjustbox}
\usepackage{multirow}
\usepackage{float}
\usepackage{makecell}

\newcommand\blfootnote[1]{%
  \begingroup
  \renewcommand\thefootnote{}\footnote{#1}%
  \addtocounter{footnote}{-1}%
  \endgroup
}

\def\cvprPaperID{11286} 
\def\confName{CVPR}
\def\confYear{2022}

\input{preamble}
\begin{document}
\raggedbottom
\input{sec/0_metadata}
\maketitle
\input{sec/0_abstract}
\input{sec/1_introduction}

\input{sec/2_related}

\input{sec/3_method}

\input{sec/4_results}
\input{sec/5_conclusions}
{
    \small
    \bibliographystyle{ieee_fullname}
    \bibliography{macros,main}
}



\end{document}

%% file: preamble.tex
\usepackage{overpic}
\usepackage{enumitem} 
\usepackage{overpic} 
\usepackage{color}

\definecolor{turquoise}{cmyk}{0.65,0,0.1,0.3}
\definecolor{purple}{rgb}{0.65,0,0.65}
\definecolor{dark_green}{rgb}{0, 0.5, 0}
\definecolor{orange}{rgb}{0.8, 0.6, 0.2}
\definecolor{red}{rgb}{0.8, 0.2, 0.2}
\definecolor{darkred}{rgb}{0.6, 0.1, 0.05}
\definecolor{blueish}{rgb}{0.0, 0.3, .6}
\definecolor{light_gray}{rgb}{0.7, 0.7, .7}
\definecolor{pink}{rgb}{1, 0, 1}
\definecolor{greyblue}{rgb}{0.25, 0.25, 1}

\usepackage{blindtext}

\renewcommand{\paragraph}[1]{\vspace{1em}\noindent\textbf{#1}.}

%% file: sec/0_metadata.tex
\newcommand*{\affmark}[1][*]{\textsuperscript{#1}}

\title{Self-Supervised Anomaly Detection by Self-Distillation and Negative Sampling}

\author{Nima Rafiee\affmark[1] \and Rahil Gholamipoorfard\affmark[1] \and Nikolas Adaloglou\affmark[1] \and Simon Jaxy\affmark[1] \and Julius Ramakers\affmark[1] \and Markus Kollmann\affmark[1,2] \\ 
\affmark[1]Department of Computer Science, Heinrich Heine University, D-40225 Dusseldorf \\ 
\affmark[2]Department of Biology, Heinrich Heine University, D-40225 Dusseldorf;\\
{\tt\small \{rafiee,rahil.gholamipoorfard,nikolaos.adaloglou,simon.jaxy,ramakers,kollmann\}@hhu.de}}


%% file: sec/0_abstract.tex
\begin{abstract}
Detecting whether examples belong to a given in-distribution or are Out-Of-Distribution (OOD) requires identifying features specific to the in-distribution. In the absence of labels, these features can be learned by self-supervised techniques under the generic assumption that the most abstract features are those which are statistically most over-represented in comparison to other distributions from the same domain. In this work, we show that self-distillation of the in-distribution training set together with contrasting against negative examples derived from shifting transformation of auxiliary data strongly improves OOD detection. We find that this improvement depends on how the negative samples are generated. In particular, we observe that by leveraging negative samples, which keep the statistics of low-level features while changing the high-level semantics, higher average detection performance is obtained. Furthermore, good negative sampling strategies can be identified from the sensitivity of the OOD detection score. The efficiency of our approach is demonstrated across a diverse range of OOD detection problems, setting new benchmarks for unsupervised OOD detection in the visual domain.

\end{abstract}

%% file: sec/1_introduction.tex
\section{Introduction}
\label{sec:intro}

\blfootnote{Under review}OOD detection or anomaly detection is the problem of deciding whether a given test sample is drawn from the same in-distribution as a given training set or belongs to an alternative distribution. Many real-world applications require highly accurate OOD detection for secure deployment, such as in medical diagnosis. Despite the advances in deep learning, neural network estimators can generate systematic errors for test examples that are far from the training set \cite{Nguyenieee}. For example, it has been shown that Deep Neural Networks (DNNs) with ReLU activation functions can make false predictions for OOD samples with arbitrarily high confidence \cite{hein2019relu}.\@ 

\input{fig/teaser}

A major challenge in OOD detection is the case where the features of outlier examples are statistically close to the features of in-distribution examples, which is frequently the case for natural images. In particular, it has been shown that deep density estimators like Variational Autoencoders (VAEs) \cite{VAE}, PixelCNNs \cite{OordKK16}, and normalising flow models \cite{rezende2016variational} can on average assign higher likelihood to OOD examples than to examples from the in-distribution \cite{nalisnick2019deep}. This surprising finding can be partially attributed to an inductive bias from upweighting local pixel correlations as a consequence of using convolutional neural networks.

A challenging scenario of anomaly detection is near OOD detection \cite{contrastano_winkens}, where the OOD distribution samples are statistically very similar to the in-distribution. A particular challenging OOD detection task is given by CIFAR$100$ \cite{krizhevsky2009learning} as in-distribution and CIFAR$10$ \cite{krizhevsky2009learning} as OOD, where the larger number of classes in CIFAR$100$ make it harder to identify features that are specific for the in-distribution. Another aspect is that OOD detection becomes more challenging if there exists a substantial class overlap between the in-distribution and the out-distribution. For instance, CIFAR$10$ and STL$10$ \cite{pmlr-v15-coates11a} share $7$ out of $10$ of their classes. Finally, there are cases where the in and out distributions are not closely related, which we refer to as far OOD. 

State-Of-The-Art (SOTA) performance has been obtained for the CIFAR$100$/CIFAR$10$ near OOD detection task, using pretrained classification models using ImageNet-21K \cite{supervised-ood-sota}. However, as CIFAR$100$ and CIFAR$10$ share many of their classes with ImageNet but the classes among themselves are mutually exclusive, the pretrained model effectively solves the OOD detection problem for this special case. The advantage of using pretrained models as OOD detectors drops if there is no class overlap with the OOD test set, such as for SVHN \cite{supervised-ood-sota}. Moreover, such massive annotated datasets with sufficient class overlap are rarely available, while unlabeled data are often widely accessible. It has also been argued that the image-level supervision may reduce the rich visual signal contained in an image to a single concept \cite{caron2021emerging,imagenet}. 

To overcome these limitations, a plethora of self-supervised pretext tasks have been proposed that provide a richer learning signal that enables abstract feature learning \cite{chen2020simple,caron2020unsupervised,he2020momentum}. These advancements in self-supervised learning have shown remarkable results on unsupervised anomaly detection \cite{csi,sehwag2021ssd,contrastano_winkens} by solely relying on the in-distribution data. 

More recently, it has been suggested to include dataset-specific augmentations that shift the in-distribution -- so-called negative samples. The core idea behind using shifting transformations is to concentrate the learned representation in feature space. This can result in a more conservative decision boundary for the in-distribution \cite{HendrycksMD19}. However, in-distribution shifting requires dataset-specific prior knowledge \cite{mohseni2021shifting}. Therefore, a bad choice of augmentations may result in rejecting the in-distribution test samples, which reduces the OOD detection performance. 

On the model side, Vision Transformers (ViT) \cite{vit} have been established in many computer vision tasks, such as image classification, and semantic segmentation \cite{xie2105segformer}. ViTs are capable of capturing long-range correlations, which is crucial for learning high-level semantics. Specifically, robust representations can be generated from ViTs by formulating a label-free self-distillation task (DINO \cite{caron2021emerging}). The DINO objective aims to map different augmentations of the same image to the same ``soft'' class. The learned features of DINO have been shown to contain explicit information about the image semantics. 

In this paper, we propose an improved version of the DINO \cite{caron2021emerging} framework in the context of OOD detection. The main contributions of this work are summarized as follows:
\begin{itemize}
 \item We propose a general methodology that leverages unlabelled data for OOD detection.
 \item We provide strategies on how negative samples can be generated in a systematic way by using the score for rejecting in-distribution test examples as OOD as a sensitivity measure. 
 \item We introduce an auxiliary loss that encourages negative samples to be uniformly assigned to the existing in-distribution soft-classes.
 \item Finally, we show that the proposed framework does not only improve OOD detection performance but also improves representation learning for the in-distribution, as measured by the K-Nearest Neighbour (K-NN) accuracy. 
\end{itemize}


%% file: fig/teaser.tex
\begin{figure}[t]
\begin{center}

\includegraphics[width=.99\columnwidth]{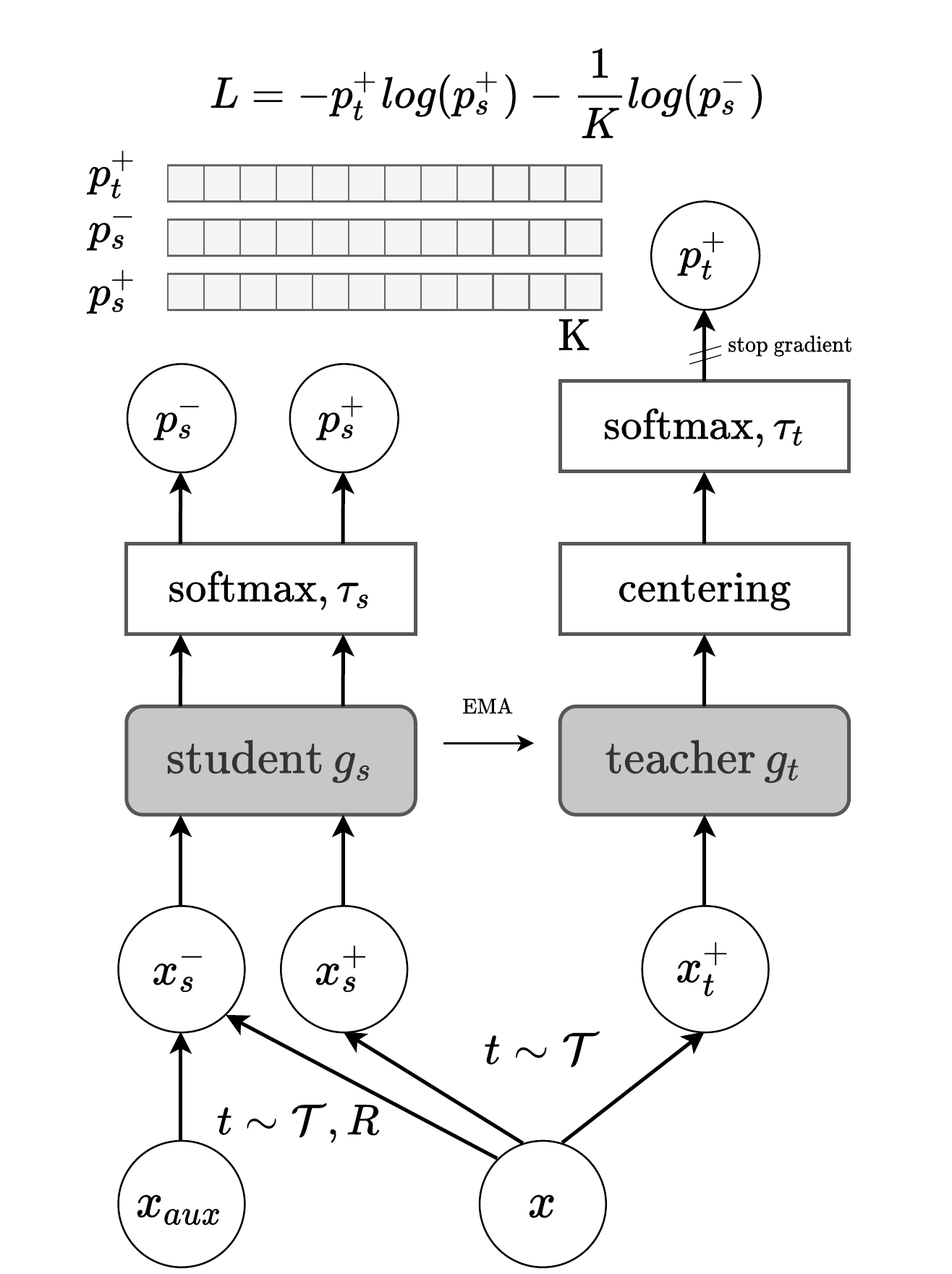}
\end{center}
\caption{An overview of the proposed contrastive self-distillation framework, consisting of student and teacher networks, $g_s$ and $g_t$, that map two random transformations of the same image, $x_s^{+}\sim\mathcal{T}(x)$ and $x_t^{+} \sim\mathcal{T}(x)$ to the same class. Negative views, $x^{-}$, arise from first applying a shifting transformation $R$, such as random rotation, followed by $\mathcal{T}$ to either an in-distribution image $x$ or an auxiliary image $x_{aux}$.}
\label{fig:teaser}
\end{figure}

%% file: sec/2_related.tex
\section{Related works}
\label{sec:related}
\noindent\textbf{Supervised OOD detection methods}. In-distribution classification accuracy is highly correlated with OOD performance \cite{fort2021exploring}. This motivated supervised OOD detection approaches to learn representations from classification networks. This can be achieved by directly training a classifier on the in-distribution or by pretraining on a larger dataset.\@ Hendrycks et al. \cite{HendrycksG17} used Maximum Softmax Probabilities (MSP) to discriminate between OOD and in-distribution samples. In-distribution classifiers have been improved by introducing additional training tricks and strategies.\@ In \cite{LiangLS18}, the authors demonstrated that the MSP performance can be increased by using a temperature parameter. In the same direction, various works focused on forming alternative loss functions \cite{LeeLLS18, VyasJZDKW18} or auxiliary objectives \cite{DeVries,HendrycksMKS19,MohseniPYW20} to learn a robust representation from the in-distribution classification.

Fine-tuning pretrained transformers \cite{vaswani2017attention} has shown promising OOD scores.\@ Hendrycks\@ et\@ al. \cite{HendrycksLWDKS20} showed that transformers are more robust to detecting outliers when pretrained on larger and more diverse natural language datasets.\@ Similarly, in computer vision, Koner et\@ al. \cite{OODformer} leveraged the contextualization capabilities of pretrained ViTs by exploiting the global image context.\@ Fort et al. \cite{fort2021exploring} fine-tuned ViTs for anomaly detection that were pretrained on ImageNet-$21$K.\@ While fine-tuning on the in-distribution,\@ they further highlighted the few-shot OOD performance of such models.\@ These kinds of large-scale pretrained models heavily rely on the classes of the pretraining dataset, which often include classes from both the in and out distribution. Hence, supervised pretraining can form a good boundary for OOD detection. However, supervised pretraining imposes two limitations for anomaly detection: a) the pretraining dataset should share labels with both distributions (in and out), and b) impeded OOD performance is observed when the distributions have overlapping classes.

Several recently developed methods \cite{LeeLLS18,Winkens2020,LG020} used annotated data to learn an intermediate representation on which a density distribution can be fitted to compute the likelihood of OOD examples.\@ In \cite{ShalevAK18}, the authors used multiple regression functions to build a robust classifier in order to identify OOD inputs, while in \cite{MasanaRSWL18} the authors used metric learning to learn an embedding where samples from the same in–distribution class form clusters.\@ Representations can be further enhanced by combining image-supervision with contrastive learning \cite{Winkens2020}. Supervised contrastive learning \cite{khosla2020supervised, ChuangRL0J20} has also been successfully applied for anomaly detection. In \cite{ChoSL21}, the authors showed an alternative way of leveraging labels by creating class-conditional masks for contrastive learning.\@ This task-specific variant of supervised contrastive learning shaped more clear boundaries between in-distribution classes, which is more befitted for OOD detection.

Mohseni et al. \cite{mohseni2021shifting} recently presented a 2-step method that initially learns how to weight the in-distribution transformations based on a supervised objective.\@ Then, the selected shifting transformations are applied in a self-supervised setup for OOD detection.\@ Still, human-level supervision is required to learn the best shifting transformations for each training dataset. In \textit{Geometric} \cite{HendrycksMKS19}, Hendrycks et al. defined a self-supervised task to predict geometric transformations to improve the robustness and uncertainty of deep learning models.\@ They further improved their self-supervised technique with supervision through outlier exposure, encouraging the network to uniformly distribute OOD samples among in-distribution classes.

\noindent\textbf{Unsupervised OOD detection methods}. Existing label-free OOD detection approaches can be separated in: a) density-based \cite{Likelihood_Ratios, Eric19, SerraAGSNL20}, b) reconstruction-based \cite{PidhorskyiAD18,ZongSMCLCC18}, and c) self-supervised learning \cite{GolanE18, HendrycksMKS19} ones. Density-based methods aim to fit a probability distribution such as Gaussian on the training data and then use it for OOD detection. Reconstruction-based methods assume that the network would generalize less for unseen OOD samples.\@ Deep generative models like VAEs were widely used for OOD detection \cite{An2015VariationalAB}, motivated by the idea that VAEs \cite{VAE} cannot reconstruct OOD samples fairly well. Meanwhile, recent studies \cite{nalisnick2019deep} revealed that probabilistic generative models can fail to distinguish between training data and OOD inputs.\@ To address this issue,\@ some efficient OOD scores were proposed based on likelihood \cite{XiaoYA20, Likelihood_Ratios, Eric19}.\@ Schirrmeister et al.\cite{SchirrmeisterZB20} leveraged the hierarchical view of distributions to propose a likelihood-based anomaly detection method.\@ More precisely, they train two identical generative architectures, one trained on the in-distribution and one on a more general distribution.

Self-supervised methods have recently shown that adopting pretext tasks results in learning general data representations \cite{DosovitskiyFSRB16} for OOD detection. Choi\@ et\@ al.\@ \cite{ChoiC20} used blurred data as adversarial examples to discriminate the training data from their blurred versions. In \cite{GolanE18}, a multi-class image classifier is trained to discriminate geometric transformations. The OOD images were then detected, by comparing the softmax probabilities of their transformed instances against train data. To extend the aforementioned method to non-imaging data, Bergman et al.\@\cite{BergmanH20} extended the set of transformations to affine transformations. 

In \textit{CSI} \cite{csi}, Tack et al. leverage shifting data transformations in contrastive learning for OOD detection, combined with an auxiliary task that predicts which shifting transformation was applied to a given input. They also demonstrated how an OOD score function can utilize contrastive representations. In \textit{SSD} \cite{sehwag2021ssd}, the authors further improved contrastive self-supervised training by developing a cluster-conditioned OOD detection method
in the feature space.

\noindent\textbf{Outlier Exposure (OE)}.\@ OE leverages auxiliary data that are utterly disjoint from the OOD data \cite{HendrycksMD19}.\@ Furthermore,\@ OE assumes that the provided auxiliary samples are always OOD.\@ To guarantee this,\@ human supervision is necessary to remove the overlap between auxiliary and in-distribution.\@ OE has been successfully applied to training classifiers, by enforcing the auxiliary samples to be equally distributed among the in-distribution classes.\@ Inspired by \cite{HendrycksMD19}, we attempt to teach the network better representations for OOD detection by incorporating auxiliary data into a self-distillation soft-labeling framework.

Finally, since the proposed method does not require labels, there is no information on whether the in-distribution data are meaningfully similar to the auxiliary ones. In this aspect, this work is different from OE, as it only requires the in-distribution to be sufficiently statistically underrepresented. To ensure the latter, an additional transformation $R$ is applied on the auxiliary data.

%% file: sec/3_method.tex
\section{Proposed Method}
\input{tab/sota}
\subsection{The vanilla DINO framework}

The DINO framework uses two identical networks $g_s = g(x \vert \theta_{s})$ and $g_t = g(x \vert \theta_{t})$ called student and teacher, which differ by their sets of parameters $\theta_{s}$ and $\theta_{t}$, respectively.\@ For each transformed input image $x$,\@ both networks produce $K$-dimensional output vectors, where $K$ is the number of soft-classes. Both outputs enter a temperature-scaled softmax functions $p_t = softmax(g_t,\tau_t)$ and $p_s = softmax(g_s,\tau_s)$ defined by:
\begin{equation}
p^i(x)=\frac{\exp \left(g^i(x) / \tau\right)}{\sum_{k=1}^{K} \exp \left(g^k({x}) / \tau\right)},
\label{sfmax}
\end{equation}
where $p^i(x)$ is the probability of $x$ falling in soft-class $i$ and $\tau_s, \tau_t$ are the student and teacher temperatures. In contrast to knowledge distillation methods, the teacher is built from previous training iterations of the student network. To do so, the gradients are back-propagated only through the student network and the teacher parameters are updated with the Exponential Moving Average (EMA) of the student parameters  

\begin{equation}
{\theta_{t}} \leftarrow m {\theta_{t}} +(1-m) {\theta_{s}},
\end{equation}

where $0\leq m \leq 1$ is a momentum parameter.\@ For $\tau_t < \tau_s$, the training objective is given by the cross entropy loss for two non-identical transformations $x'',x^{\prime}$ of an image $x$ drawn from the in-distribution training set $\mathcal{D}_{train}^{in}$

\begin{equation}
  \mathcal{L}_{pos} = 
  - \sum_{x''\in G}\,\sum_{\substack{x^{\prime}\in V \\ x^{\prime}\neq x''}} p_t(x'') \log(p_s(x^{\prime})).
  \label{CE}
\end{equation}

Additionally,\@ DINO uses the multi-crop strategy \cite{caron2020unsupervised},\@ wherein $M$ global views $G=\{x_1^g,...,x_M^g\}$ and $N$ local views,\@ $L=\{x_1^l, \cdot\cdot\cdot,x_N^l\}$, are generated based on a set of transformations $\mathcal{T}$, e.g. crop and resize, horizontal flip, Gaussian blur, and color jitter. Global views are crops that occupy a larger region of the image (e.g. $\geq 40\%$) while local views cover small parts of the image (e.g. $\leq 40\%$).\@ All $V=G\cup L$ views are passed through the student network,\@ while the teacher has only access to the global views such that local-to-global correspondences are enforced.\@ The trained teacher network is used for evaluation.

\subsection{Negative samples}
The learning objective (\cref{CE}) assigns two transformed views of an image to the same soft-class. The applied transformations $\mathcal{T}$ are chosen to be sufficiently strong and diverse, such that the generated images generalise well over the training set but keep the semantics of the image they were derived from. The transformations are designed to learn higher-level features such as labels that represent semantic information and avoid learning lower-level features, such as edges or the color statistics over pixels \cite{chen2020simple}. The quality of the learned representation can be quantified by evaluating the K-NN accuracy for an in-distribution test set $\mathcal{D}_{test}^{in}$, using as higher-level feature vector an activity map of the network near the last layer. For OOD detection, the feature vector representation should be enriched by in-distribution-specific features and depleted by features that frequently appear in other distributions from the same domain. This can be achieved by designing a negative distribution $D_{neg}$ that keeps most of the low-level features of the in-distribution but changes the high-level semantics. 

For example, a negative distribution for natural images can be realised by additionally rotating in-distribution images or images from a related auxiliary distribution by $r\sim R=\mathcal{U}(\{90^{\circ},180^{\circ},270^{\circ} \})$, where $\mathcal{U}$ is the uniform distribution. It has been shown that using rotation as an additional positive transformation degrades the performance in the contrastive learning setup, where the objective is to maximize the mutual information between positive examples \cite{chen2020simple}. Motivated by this, authors in \cite{csi} report a performance gain for OOD detection by using rotation to generate negative examples. 

\subsection{Auxiliary objective}
In addition to the self-distillation objective \cref{CE} we define an auxiliary task to encourage the student to have a uniform softmax response for negative examples.\@ This task can be realised by a similar objective as \cref{CE} but with changed temperature $\tau_{t} \rightarrow \infty$ and transformations $\mathcal{T}$ applied to examples $x$ from the negative set $\mathcal{D}_{neg}$, defined as:
\begin{equation}
  \mathcal{L}_{neg}=-\frac{1}{K} \sum_{x^{\prime}\in V}\log p_s(x').
  \label{L_aux}
\end{equation}
The total loss of our proposed method is defined by a linear combination of the two objectives
\begin{equation}
  \mathcal{L}_{total} = \mathcal{L}_{pos} + \lambda \mathcal{L}_{neg},
  \label{L_total}
\end{equation}
where $\lambda > 0$ is a balancing hyperparameter.\@


%% file: tab/sota.tex
\begin{table*}[t]
\caption{AUROC scores for OOD detection without label supervision.}
\label{tabl1_sota}
\begin{center}
\begin{tabularx}{\textwidth}{ll c*{6}{Y}}
\hline
&&\multicolumn{6}{c}{OOD Detection AUROC ($\%$)}\\
\cline{3-8}
& & & & & &\multicolumn{2}{c}{\textbf{Ours}}\\
\cline{7-8}
{$\mathcal{D}_{train}^{in}$} & {$\mathcal{D}_{test}^{out}$} & Geometric$^*$\cite{HendrycksMKS19} & SSD\cite{sehwag2021ssd} & CSI\cite{csi} &  MTL$^{\dagger}$\cite{mohseni2021shifting} &\parbox[t]{1.5cm}{Rot.\\ImgN} & \parbox[t]{1.5cm}{Combined} \\
\hline
\multirow{8}{*}{\rotatebox{90}{CIFAR$10$}} & CIFAR$100$ &
                   $91.91$ & $90.63$ &  $89.20$ 
                    & $93.24$ & $92.51$ & $\mathbf{94.20}$ \\

& SVHN & 
                   $97.96$ & $99.62$  &$99.80$ 
                   & $\mathbf{99.92}$ & $99.69$  & $\mathbf{99.92}$ \\ 

& ImageNet$30$ & 
                   $-$ & $90.20$ &  $87.92$ & $-$ & $\mathbf{94.16}$ & 
                   $93.40$ \\ 

& TinyImageNet & 
                   $92.06$ & $92.25$ & $92.44$ & $92.99$ & $\mathbf{96.28}$ & 
                   $95.02$ \\

& LSUN & 
                   $93.57$ & $96.51$ &  $91.60$ & $95.03$ & $\mathbf{98.08}$ & 
                   $97.52$ \\ 

& STL$10$ & 
                   $-$ &$70.28$ & $64.25$ & 
                   $-$ & $\mathbf{77.29}$ &
                   $74.34$ \\ 
 
& Places$365$ & 
                   $92.57$ & $95.21$ &  $90.18$ & $93.72$ & $\mathbf{97.14}$ & 
                   $96.01$ \\
                   

& Texture & 
                   $96.25$ & $97.61$  &$98.96$& 
                   $-$ &  $\mathbf{99.16}$ & $98.69$ \\

                   
\hline \hline
\multirow{8}{*}{\rotatebox{90}{CIFAR$100$}}
& CIFAR$10$& 
                  $74.73$ & $69.60$  &$58.87$ & $\mathbf{79.25}$ & $69.96$ & $67.63$ \\

& SVHN & 
                  $83.62$ & $94.90$ &  $96.44$ & $87.11$ & $96.00$ &
                  $\mathbf{97.17}$ \\

& ImageNet$30$ &
                  $-$ & $75.53$ & $71.82$  & $-$ & $\mathbf{84.82}$ & 
                  $75.36$ \\

& TinyImagenet & 
                  $77.56$ & $79.52$ &$79.28$ & $80.66$ & $\mathbf{81.41}$ &
                  $79.75$ \\
 
& LSUN & 
                  $71.86$ & $79.50$ & $61.83$ & $74.32$ & $\mathbf{85.03}$ &
                  $74.55$ \\

& STL$10$ & 
                  $-$ &$72.76$ & $64.26$ & $-$ & $\mathbf{79.96}$ &
                  $71.70$ \\

& Places$365$ & 
                  $74.57$ & $79.60$ & 
                  $65.48$ & $77.87$ & $\mathbf{81.67}$ &
                  $72.79$ \\

& Texture & 
                  $82.39$ & $82.90$ &  $\mathbf{87.47}$& 
                  $-$ & $80.65$ &
                  $77.33$ \\
\hline
\end{tabularx}
\end{center}
\footnotesize{$^{*}$ Requires labels for the supervised training loss. Results reported from \cite{mohseni2021shifting}.}\\
\footnotesize{$^{\dagger}$ Requires labels to select the optimal transformations.}
\end{table*}

%% file: sec/4_results.tex

\section{Experiments}
The proposed method is based on the vanilla DINO \cite{caron2021emerging} implementation\footnote{https://github.com/facebookresearch/dino}.\@ Unless otherwise specified, we use ViT-Small (ViT-S) with a patch size of $16$.\@ We use $N\!=\!8$ local views for both positives and negatives, but two global positive views and one global negative view. \@ Global views are resized to $256\times 256$ while local views to $128\times 128$.\@ The temperatures are set to $\tau_t=0.01$ and $\tau_s=0.1$.\@ In each epoch, we linearly decrease $\tau_t$ starting from $0.055$ for CIFAR$10$ and from $0.050$ for CIFAR$100$ to $0.01$ during training.\@ We set $\lambda$ to $1$ for all our experiments, and $K=4096$.

We use the $\operatorname{Adamw}$ optimizer \cite{loshchilov2018fixing} with an effective batch size of $256$.\@ The learning rate $lr$ follows the linear scaling rule  of $lr\!=\! \mathrm{lr_{base}}\times \operatorname{batch size}/256$,\@ where $\mathrm{lr_{base}}\!=\!0.004$.\@ All models are trained for $500$ epochs.\@ Experiments were conducted using $4$ NVIDIA-A$100$ GPUs with $40$GB of memory. The image augmentation pipeline $\mathcal{T}$ is based on \cite{byol,caron2021emerging}. Finally, weight decay and learning rate are scaled with a cosine scheduler. 

\input{tab/cifar10_Ablation_table}
\subsection{Datasets and negative sample variants} \label{negative-sample-strategies}
We evaluate our method on CIFAR$10$ and CIFAR$100$ as in-distribution data.\@ For auxiliary datasets, we use ImageNet \cite{imagenet} and Debiased $300$K Tiny Images (DTI) \cite{HendrycksMD19}.\@ The latter is a subset with $300$K images from \cite{80milliontiny}, where images belong to CIFAR$10$, CIFAR$100$, Places$365$ \cite{7968387}, and LSUN \cite{DBLP:journals/corr/YuZSSX15} classes are removed.\@ To avoid shortcut learning (due to different image resolutions),\@ we resize the auxiliary data to the size of the in-distribution data before applying any augmentation.\@ For OOD detection,\@ we consider common benchmark datasets, such as SVHN \cite{Netzer2011ReadingDI}, Places$365$, Texture \cite{DBLP:journals/corr/CimpoiMKMV13} and STL$10$.\@ The following cases are considered for generating negative samples:

\begin{itemize}
    \item DINO: no negatives are included ($\lambda=0$).
    \item ImgN: samples from ImageNet.
    \item DTI: samples from Debiased Tiny Images.
    \item Rot.: samples are randomly rotated by $r\sim R=\mathcal{U}(\{90^{\circ},180^{\circ},270^{\circ} \})$.
    \item Rot.360: samples are randomly rotated between $0^{\circ}$ and $360^{\circ}$ in $90^{\circ}$ steps.
    \item Perm-$N$: randomly permutes each part of the evenly partitioned image in $N$ patches.
    \item Pix. Perm: randomly shuffles all the pixels in the image.
    \item Rot. In-Dist: a random rotation $r\sim R$ is applied to the in-distribution data.
    \item Combined: both samples from Rot. In-Dist and Rot. ImageNet are used.
\end{itemize}

\subsection{Evaluation protocol for OOD detection}
The DINO network structure $g(x)$ used in this work consists of a ViT-S as backbone, which maps the input $x$ to a d-dimensional feature vector $f\in \mathbb{R}^d$, and two fully connected layers as head, which converts the features vector $f$ to a $K$-dimensional output vector that enters the softmax layer. We define an anomaly detection score, $\mathcal{S}$, for the OOD test data $\mathcal{D}_{test}^{out}$ by computing the cosine similarity between the feature vector for a test image $f_{test}$ and all features vectors $f_m$ of the in-distribution training set. Instead of taking the maximum cosine similarity as a OOD score, we opt for a temperature weighted non-linear score,
\begin{equation}
 \centering
 \mathcal{S}(x) = - \frac{1}{M}\sum_{m=1}^{M} \exp\left({\frac{1}{\tau}\cdot \frac{f_{test}^{T}f_m}{\|f_{test}\| \|f_m\|}}\right), \quad \label{eq:cosine_sim}
\end{equation}
with $\tau\!=\!0.04$ a fixed temperature and $M$ the number of in-distribution training samples.\@ The score is used to evaluate OOD performance by reporting the Area Under the Receiver Operating characteristic Curve (AUROC) between a given OOD test set and the in-distribution test set.

\subsection{Experimental results}
In \cref{tabl1_sota}, quantitative results are reported for CIFAR$10$ and CIFAR$100$ as in-distribution. We report results with ImageNet rotated samples as well as combining them with in-distribution rotated samples (Combined).
When using CIFAR$10$ as $\mathcal{D}_{train}^{in}$, the proposed method shows superior  performance in 6 out of 8 (75\%) OOD datasets compared to current SOTA self-supervised methods. Surprisingly, we even surpass hybrid methods, where self-supervised training is combined with human-labelled images. By further leveraging in-distribution negatives, we are able to surpass all other methods in CIFAR$100$ by $3.57\%$ and $0.96\%$ against self-supervised and supervised methods, respectively.\@ Therefore, we believe that near OOD performance can have a significant improvement when there is prior knowledge in terms of choosing the appropriate dataset-specific transformation to form good negative examples, applied on $\mathcal{D}_{train}^{in}$, such as rotating CIFAR$10$ images. \par
Our results are roughly consistent for CIFAR$100$ as $\mathcal{D}_{train}^{in}$. Again, we report superior performance in $6$ out of $8$ (75\%). Far OOD datasets have a substantial benefit, such as LSUN where we report a 5.53\% gain against the best self-supervised method. Still, our results on near OOD on CIFAR$10$ are on par with self-supervised methods \cite{csi}, while still lacking behind supervised methods. As illustrated in \cref{tabl1_sota}, there was no gain in near OOD when adding rotated in-distribution samples, which is justified by the fact that the provided rotations do not form as good negative samples as in CIFAR$10$. Our findings on the chosen shifting transformations are in line with \cite{mohseni2021shifting}, wherein translation is considered to be the best choice for CIFAR$100$.

In \cref{tab:contrast_dataset_cif10}, we investigate several ways to generate negative samples, as detailed in \cref{negative-sample-strategies}. It can be observed that by rotating both ImageNet and DTI with $R$, both distributions demonstrate an average performance gain of $2.63\%$ and $1.55\%$ respectively compared to no additional transformation. On the contrary, when applying the Rot.$360$ transformation on ImageNet, performance deteriorates by $1.57\%$ on average.   
\input{fig/occupied_classes}
Rotated ImageNet and DTI reached the highest gains of $10.79\%$ and $14.99\%$ in STL$10$ compared to DINO. We claim that leveraging auxiliary rotated datasets best suits cases when there is a big class overlap, such as CIFAR$10$ and STL$10$. Interestingly, rotated CIFAR$10$ samples outperform all other strategies in near OOD (CIFAR$100$). This finding further confirms that dataset-specific shifting transformations can form good near OOD boundaries when treated as negative examples. Apart from OOD detection in CIFAR$100$, Rot. In-Dist still shows inferior results compared to Rot. ImageNet and Rot. DTI.

It is worth noting that we abstain from reporting the performance of DTI in \cref{tabl1_sota}, since labels were used to form this subset of $300$K images. Nonetheless, we show that the introduced method is not specifically linked to ImageNet, but only assumes that a broad distribution of unlabelled data is available. On top of that, the reported results indicate that one can use fewer image samples than ImageNet. Finally, we report an inferior (or on par) average AUROC score when employing Pix. Perm, Perm-4, and Perm-16 against the vanilla DINO method using ImageNet as the auxiliary dataset.

\section{Discussion}
\noindent\textbf{Do negative samples lead to more condensed in-distribution representations?}
To understand the impact of the introduced negative sampling methods, we investigate how many of the $K=4096$ soft-classes are ``occupied'' by the $\mathcal{D}_{test}^{in}$ after training on CIFAR$10$. A soft-class is considered occupied if the probability assigned to  that soft-class from all test data is greater than the average soft-class probability. As depicted in \cref{fig:occupied_classes} (left), negative sampling reduces the occupied classes compared to the DINO baseline.\@ This observation is independent of how $\mathcal{D}_{neg}$ is created.\@ 
More specifically, Rot. ImageNet, Rot. DTI, and Rot. In-Dist use roughly the same number of soft-classes and achieve SOTA AUROC scores on CIFAR$100$.\@ By combining the aforementioned qualitative evaluations with \cref{tab:contrast_dataset_cif10}, we claim that by contrasting $\mathcal{D}_{train}^{in}$ against $\mathcal{D}_{neg}$ a more condensed representation can be learnt.

By incorporating additional transformations, the negative samples become more dissimilar to $\mathcal{D}_{train}^{in}$, which renders the representations to be even more condensed. Besides, the transformations $\mathcal{T}$ applied on $\mathcal{D}_{train}^{in}$ (i.e. crops, jitter) are keeping $\mathcal{D}_{train}^{in}$ and $\mathcal{D}_{test}^{in}$ close together. This finding is considered a promising research direction for future work. 

In addition, a relationship between AUROC scores on CIFAR$10$ against CIFAR$100$ and the occupied classes is highlighted in \cref{fig:occupied_classes} (right). In particular, negative sampling strategies can be evaluated by looking at $\mathcal{D}_{test}^{in}$. Nonetheless, we note that this correlation becomes weaker when comparing average AUROC percentages across all considered datasets, especially when the negative sampling strategy performs worse than the baseline. 

\input{fig/4_plots}
\noindent\textbf{Is OOD detection related to in-distribution classification?}
To answer this question,\@ we investigate if there is a relationship between the OOD detection performance and the K-NN accuracy, determined from human-generated labels.\@ To do so, we use CIFAR$10$ as $\mathcal{D}_{train}^{in}$ and CIFAR$100$ and Texture as $\mathcal{D}_{test}^{out}$, as representative cases of near OOD and far OOD respectively.\@ We find that the OOD AUROC score is positively correlated with K-NN accuracy for both near and far OOD detection (\cref{fig:scatter_plots4x}, top row).  

\noindent\textbf{How to choose good negative examples?}
Ideally, the distribution of negative examples, $D_{neg}$, should share most of the features' statistics of the in-distribution, $D^{in}$, but discrimination between negative examples and in-distribution examples should be possible in practice, e.g. by a deep neural network.\@ The statistical closeness between $D_{neg}$ and $D^{in}$ is necessary to ensure high detection performance for near ODD examples. In practice, the design of $D_{neg}$ is difficult in absence of labels and some domain knowledge is needed to avoid significant overlap with the in-distribution. In this work, we apply rotation to images of natural objects to reduce the overlap with the in-distribution.

To quantify the statistical relatedness we apply our OOD score (\cref{eq:cosine_sim}) to a test set of in-distribution examples. The degree of rejection of in-distribution examples gives us a measure about the sensitivity of the OOD score to examples that have very similar features statistics to $\mathcal{D}_{train}^{in}$. Based on this measure, we find that the combination of rotated ImageNet examples and rotated in-distribution examples (``Combined'') generates the statistically closest negative examples for CIFAR$10$ among all the $D_{neg}$ we used in experiments (\cref{fig:scatter_plots4x}). This result is confirmed in \cref{tabl1_sota} and \cref{tab:contrast_dataset_cif10}, where the near OOD detection problem CIFAR$10$/CIFAR$100$ receives the highest {AUROC} score for ``Combined''. However, using ``Combined'' as $D_{neg}$ is not the best option for the semantically far OOD detection problem CIFAR$10$/Texture (\cref{tab:contrast_dataset_cif10}). This can be explained by the resulting insensitivity of the OOD score to all the low-level features shared between the in-distribution and rotated in-distribution that may help to reject Texture examples as OOD. In this case, taking rotated ImageNet examples as negatives is a better option.


%% file: tab/cifar10_Ablation_table.tex
\begin{table*}[t]
\caption{AUROC scores for OOD Detection with CIFAR$10$ as {$\mathcal{D}^{in}_{train}$} and different $\mathcal{D}_{neg}$. ImgN denotes ImageNet samples.} 
\label{tab:contrast_dataset_cif10}
\setlength{\tabcolsep}{2pt}
\begin{center}
\begin{tabularx}{\linewidth}{c*{11}{Y}}
\hline

\multicolumn{2}{c|}{Negative Sampling:} &  \multicolumn{1}{Y|}{None} & \multicolumn{8}{c|}{Auxiliary} & \multicolumn{1}{Y}{In-Dist} \\
\hline
\parbox[t]{1.0cm}{\centering \medskip \smallskip $\mathcal{D}_{train}^{in}$}  &
\parbox[t]{1.85cm}{\centering \medskip \smallskip $\mathcal{D}_{test}^{out}$ } &  DINO  $\lambda=0$ & ImgN & Rot. ImgN & Rot.$360$ ImgN & DTI & Perm-$16$ ImgN & Perm-$4$ ImgN & \parbox[t]{1.0cm}{\centering Rot.\\DTI} & Pix. Perm. & Rot. In-Dist. \\

\hline

\multirow{11}{*}{} 
\multirow{8}{*}{\rotatebox{90}{CIFAR$10$}}

&
  \multicolumn{1}{c}{CIFAR$100$} &
                   $90.29$ & $90.46$ & $92.51$ & $88.62$ & $93.77$ & $88.32$ & $89.57$ & $93.77$ & $87.67$ & $\mathbf{93.96}$ \\

& \multicolumn{1}{c}{SVHN} &  
                   $99.38$ & $99.50$ & $99.69$ & $99.42$ & $99.86$ & $99.59$ & $99.13$ & $99.86$ & $99.62$ & $\mathbf{99.92}$  \\

& \multicolumn{1}{c}{ImageNet$30$} & 
                   $88.81$ & $89.96$ & $94.16$ & $88.95$ & $93.39$ & $89.17$ & $84.71$ & $\mathbf{96.04}$ & $87.46$ & $91.69$  \\

& \multicolumn{1}{c}{TinyImageNet} & 
                   $91.07$ & $94.14$ & $\mathbf{96.28}$ & $91.60$ & $94.53$ & $89.72$ & $91.27$ & $95.64$ & $89.39$ & $94.27$  \\                   

& \multicolumn{1}{c}{LSUN} & 
                   $92.20$ & $93.41$ & $98.08$ & $93.24$ & $98.56$ & $94.58$ & $89.32$ & $\mathbf{99.12}$ & $93.33$ & $94.93$  \\                   

& \multicolumn{1}{c}{STL$10$} & 
                   $66.50$ & $77.65$ & $77.29$ & $72.41$ & $72.01$ & $69.22$ & $68.81$ & $\mathbf{81.49}$ & $68.55$ & $69.11$  \\

& \multicolumn{1}{c}{Places$365$} & 
                   $91.28$ & $93.12$ & $97.14$ & $92.58$ & $97.03$ & $92.77$ & $87.63$ & $\mathbf{98.12}$ & $91.89$ & $93.53$  \\

& \multicolumn{1}{c}{Texture} & 
                   $96.21$ & $95.01$ & $\mathbf{99.16}$ & $93.93$ & $97.55$ & $93.38$ & $89.86$ & $95.11$ & $93.08$ & $98.29$  \\
\hline
& \multicolumn{1}{c}{Average} & 	
                   $89.47$ & $91.66$ & $94.29$ & $90.09$ & $93.34$ &
                   $89.59$ & $87.54$ & $\mathbf{94.89}$ & $88.87$ & $91.96$ \\
            
\hline
\end{tabularx}

\end{center}
\end{table*}

%% file: fig/occupied_classes.tex
\begin{figure*}[t]
\begin{center}

\includegraphics[width=1.99\columnwidth]{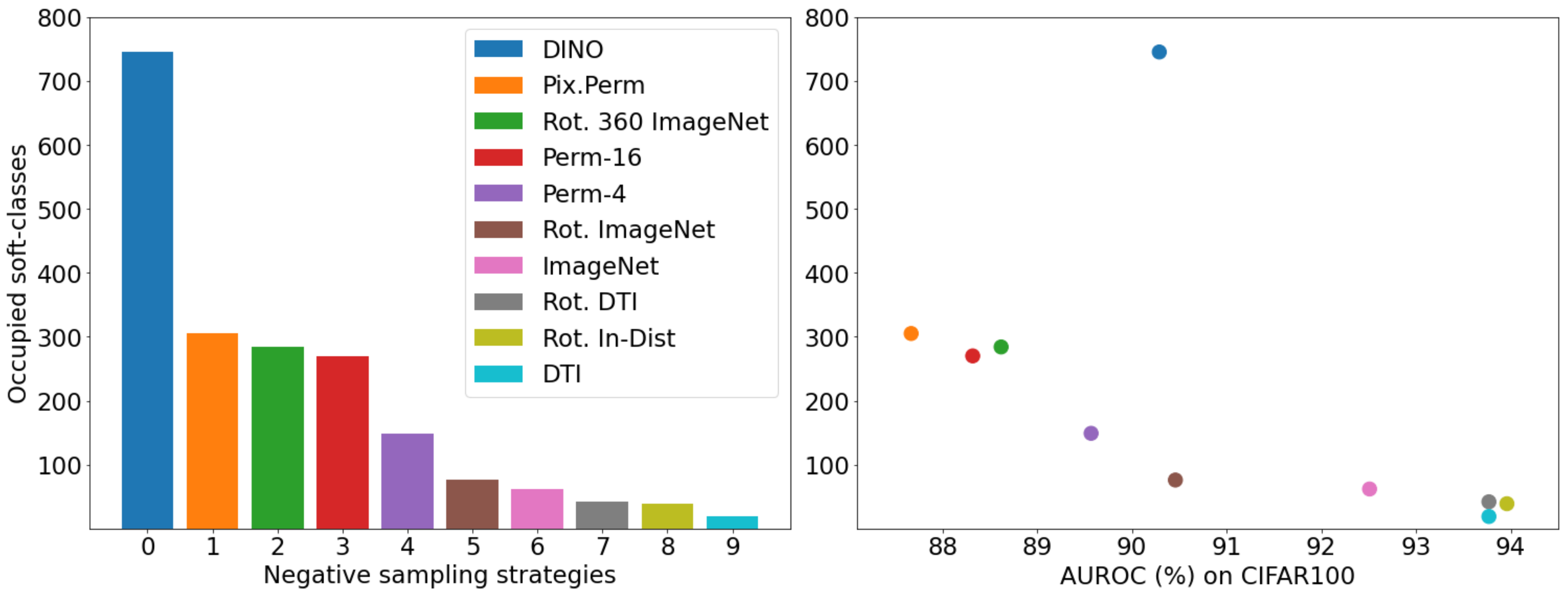}
\end{center}
\caption{We define a soft-class as ``occupied'' if the probability assigned to that soft-class is greater than the average probability of all $K$ soft-classes. Colors indicate multiple $\mathcal{D}_{neg}$ and are shared within the two plots. The teacher network $g_t$ is used to generate $p_t$ from $\mathcal{D}_{test}^{in}$. Training is performed on CIFAR$10$. \textbf{Left}: $\mathcal{D}_{test}^{in}$ occupy less soft-classes with negative sampling compared to the DINO baseline. \textbf{Right}: relationship of occupied soft-classes with respect to AUROC score in CIFAR$100$.}
\label{fig:occupied_classes}
\end{figure*}

%% file: fig/4_plots.tex
\begin{figure*}[t]
\begin{center}
\includegraphics[width=1.99\columnwidth]{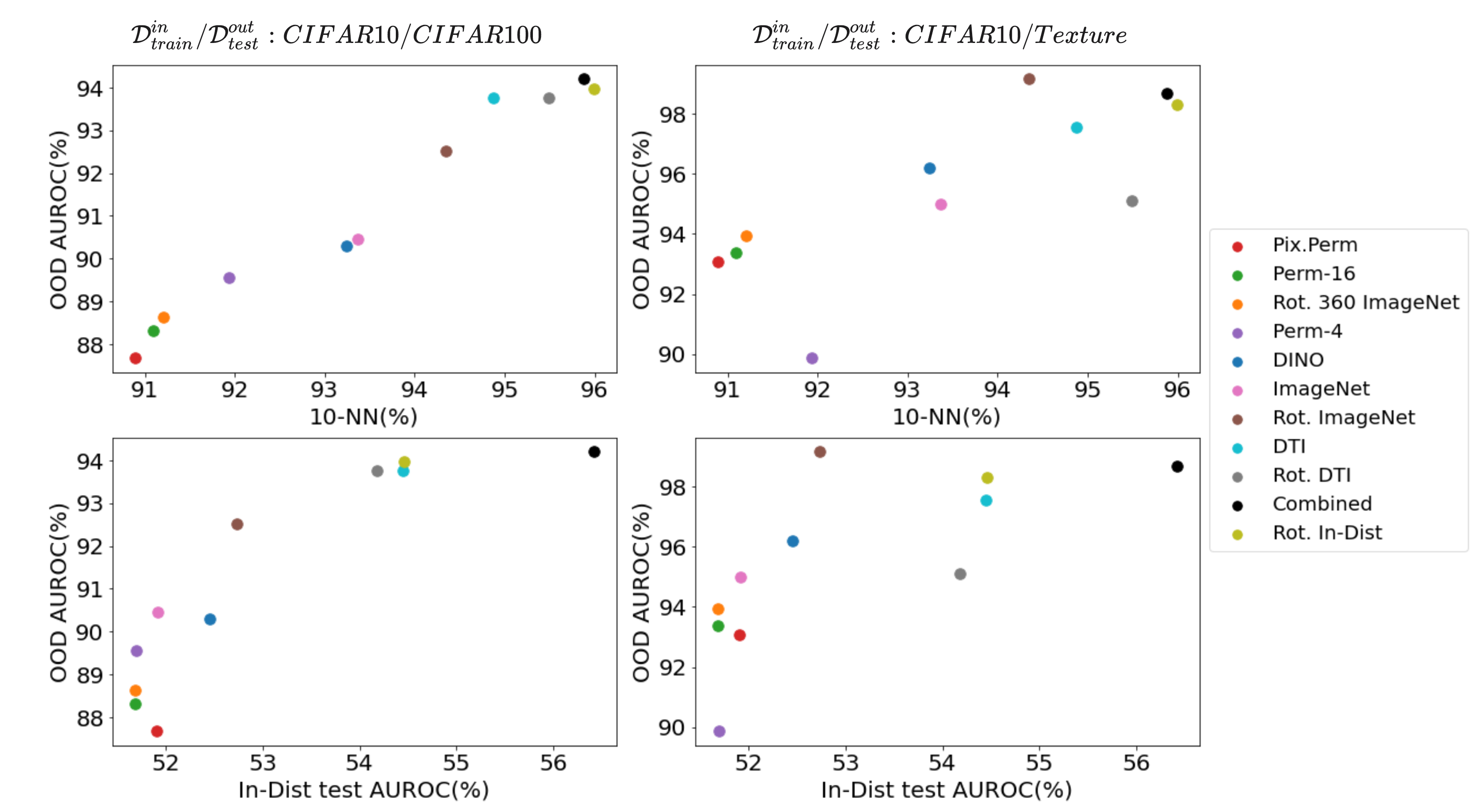}
\end{center}
\caption{We evaluate different models trained on CIFAR$10$ for two OOD datasets, CIFAR$100$ (left column) and Texture (right column). In each plot, points indicate different negative sampling strategies (colors are shared).\@ \textbf{Top row}: correlation between OOD detection AUROC and K-NN accuracy on $\mathcal{D}_{test}^{in}$.\@ \textbf{Bottom row}: correlation between OOD detection AUROC  and AUROC score of $\mathcal{D}_{train}^{in}$ vs. $\mathcal{D}_{test}^{in}$. We observe models with higher sensitivity to detect $\mathcal{D}_{test}^{in}$ as outliers have higher OOD detection performance.}
\label{fig:scatter_plots4x}
\end{figure*}


%% file: sec/5_conclusions.tex
 
\section{Conclusion}
In this work, we presented a new general method for self-supervised OOD detection.\@ We demonstrated how self-distillation can be extended to account for positive and negative examples by introducing an auxiliary objective.\@ The proposed objective introduces a form of contrastive learning, which pushes negative samples to be uniformly distributed among the existing in-distribution soft-classes.\@ Additionally, we thoroughly studied how negative samples can be generated by comparing multiple variations, based on two auxiliary datasets.\@ The different negative sampling approaches were compared in terms of OOD detection performance, as well as in terms of their impact on the in-distribution classification.\@ Insights regarding choosing transformations with respect to near and far OOD were provided.\@ The proposed method outperforms current SOTA for self-supervised OOD detection methods in the majority of OOD benchmark datasets for both CIFAR$10$ and CIFAR$100$ as $\mathcal{D}_{train}^{in}$.\@ We hope that the provided insights of our analysis will shed light on how to choose negative samples in more challenging vision domains.    